\documentclass[letterpaper, 10 pt, conference]{ieeeconf} 
\IEEEoverridecommandlockouts                             
\usepackage{cite}
\usepackage{graphicx}
\usepackage[caption=false,font=footnotesize]{subfig}
\usepackage{amsmath}
\usepackage{url}
\usepackage{algorithmic}
\usepackage{float}
\usepackage{threeparttable}
\usepackage{overpic}
\usepackage{psfrag}
\usepackage{hyperref}

\usepackage[dvipsnames]{xcolor}

\begin{document}

\bstctlcite{IEEEexample:BSTcontrol}

\title{\LARGE \bf Tenodesis Grasp Emulator:
\\Kinematic Assessment of Wrist-Driven Orthotic Control}



\author{Erin Y. Chang*, Raghid Mardini, Andrew I. W. McPherson, Yuri Gloumakov, and Hannah S. Stuart
    \thanks{E.Y. Chang, R. Mardini, A.I.W. McPherson, Y. Gloumakov, and H.S. Stuart are with the Embodied Dexterity Group, Dept. of Mechanical Engineering, University of California Berkeley, Berkeley, CA, USA.}
    \thanks{* Corresponding author {\tt\small erin.chang@berkeley.edu}}
    \thanks{This paper has a supplemental video associated with it.} 
}

\maketitle

\thispagestyle{empty}
\pagestyle{empty}

\begin{abstract}
Wrist-driven orthotics have been designed to assist people with C6-7 spinal cord injury, however, the kinematic constraint imposed by such a control strategy can impede mobility and lead to abnormal body motion. This study characterizes body compensation using the novel Tenodesis Grasp Emulator, an adaptor orthotic that allows for the investigation of tenodesis grasping in subjects with unimpaired hand function. Subjects perform a series of grasp and release tasks in order to compare normal (test control) and constrained wrist-driven modes, showing significant compensation as a result of the constraint. A motor-augmented mode is also compared against traditional wrist-driven operation, to explore the potential role of hybrid human-robot control. We find that both the passive wrist-driven and motor-augmented modes fulfill different roles throughout various tasks tested. Thus, we conclude that a flexible control scheme that can alter intervention based on the task at hand holds the potential to reduce compensation in future work.

\end{abstract}

\section{Introduction}
\label{sec:intro}

\begin{figure}[t]
    \centering
    \vspace{2mm}
    \includegraphics[width=1\linewidth]{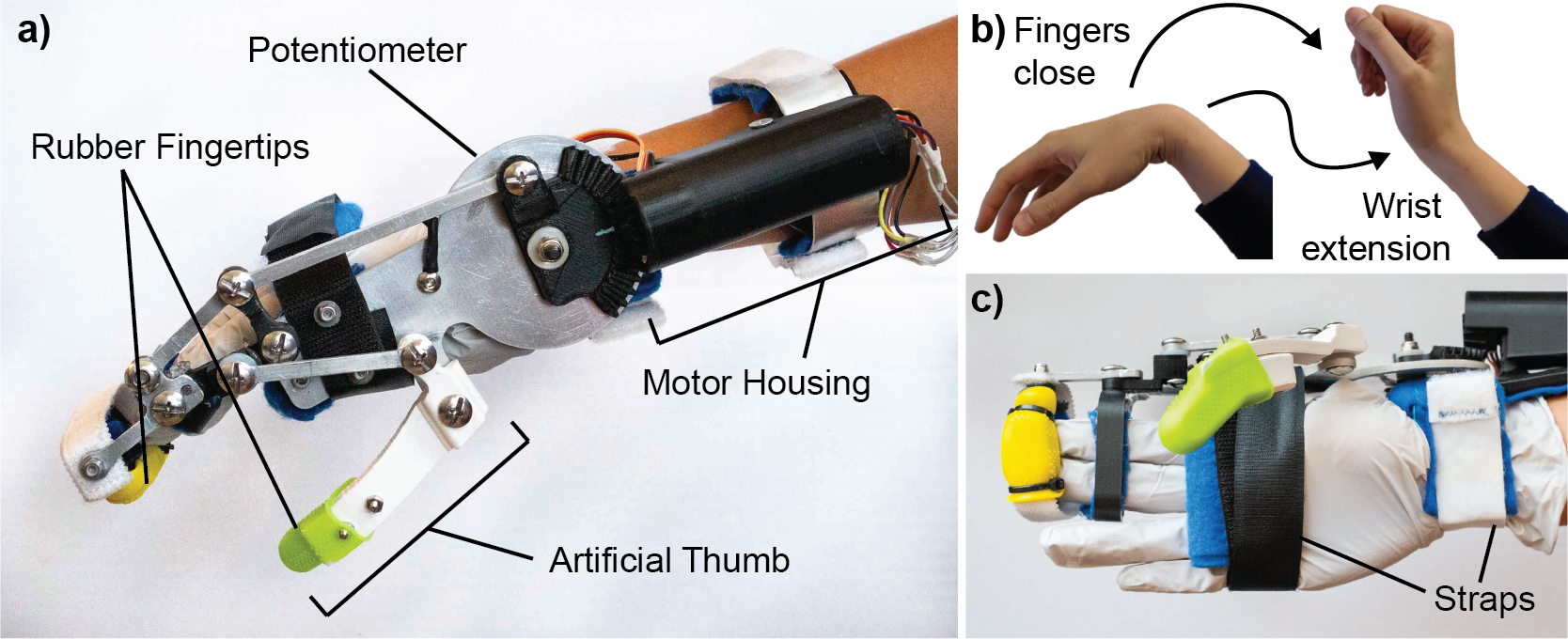}
    \vspace{-5mm}
    \caption{(a) Tenodesis Grasp Emulator (TGE) worn with the thumb tucked against the palm and secured with straps. (b) Tenodesis grasp motion performed by extending the wrist to passively close the fingers (adapted from \cite{McPherson2020Motor-AugmentedInjury}). (c) Underside of TGE, where the thumb is strapped to the palm.}
    \vspace{-5mm}
    \label{fig:testbed}
\end{figure}

The human hand plays a central role in everyday life, interacting with the environment, manipulating objects, and non-verbally conveying information. Without normative hand functionality, individuals may have difficulty performing activities of daily living (ADLs) and must rely on assistance from others or develop new methods to regain lost dexterity. In the United States alone, as many as 46,000 people live with tetraplegia due to cervical-level spinal cord injury (SCI)
\cite{NSCISC2021}, which severely reduces hand functionality \cite{bromley2006tetraplegia}. Though their arms and hands are greatly affected, individuals with SCI at the C6-7 level retain the ability to extend the wrist and often learn alternative ways to perform prehensile tasks. One common method is the tenodesis grasping technique (Fig. \ref{fig:testbed}b), a passive motion that couples finger flexion with wrist extension to produce a gentle grasp \cite{harvey2010effect, mateo2013kinematic}.
Individuals with tetraplegia report arm and hand function as their highest priority for improving quality of life \cite{KD2004TargetingPopulation}. 

Wearable technology holds great potential to enhance quality of life by restoring dexterous manipulation abilities. One device prescribed to individuals with SCI is the wrist-driven orthosis (WDO), designed to improve existing tenodesis grasping capability with passive mechanical linkages \cite{Shepherd1991}. Body-powered devices like the WDO offer the wearer a kinesthetic understanding of grasping \cite{abbott2021}, a property commonly presumed to improve device embodiment. 
However, wheelchair users with varied levels of trunk and arm mobility frequently abandon the WDO in favor of a set of more specialized tools, static braces, or no device at all \cite{Phillips1993PredictorsAbandonment}. Previously, we presented a novel device design with a motorized input on the typically body-powered WDO (MWDO) \cite{McPherson2020Motor-AugmentedInjury}. We hypothesized that a motor-augmented control method could improve the efficacy of the device and dexterity of individuals with SCI during tenodesis grasping and in this work, we follow up with an evaluation on users.



Prior work suggests that constraining wrist position during reach-to-grasp results in excessive compensatory movements \cite{Kaneishi2019HybridInjuries,Montagnani2015IsProstheses,Spiers2018ExaminingWorkspace}. We hypothesize that coupled hand and wrist function in tenodesis grasping likewise induces atypical and exaggerated upper body motions and present the first study in characterizing this kinematic constraint. 
We also investigate the consequences on body motion of two tenodesis grasp-based control strategies, resembling the WDO and MWDO devices, to determine whether an additional motorized input can reduce compensation in some instances. Ultimately, this work aims to inform the realization of new hybrid interventions that harness both body- and motor-power to provide responsive assistance to reduce unnecessary body motions and better address ADL use.

Recruiting users from vulnerable populations can result in few research subjects, 
often reducing research outcomes to case studies and delaying the development of new prototypes.
Therefore, adaptor devices have been utilized in fields such as the development of upper-limb prostheses \cite{Montagnani2015IsProstheses,GloumakovTrajectoryDevices,Carey2009KinematicActivities}, yet no such device exists in the context of tenodesis grasping for individuals with SCI. We introduce a novel investigational exoskeleton, the Tenodesis Grasp Emulator (TGE) shown in Fig. \ref{fig:testbed}a, that enables the study of tenodesis grasp in normative subjects, or individuals without SCI.

We proceed to introduce the TGE device design in Section \ref{sec:design}, followed by the details of the human subjects trial and analysis method in Section \ref{sec:methods}. In Sections \ref{sec:results} and \ref{sec:discussion}, we describe the kinematic findings from both passive and motor-assisted tenodesis grasping, then end with our conclusions and intentions for future work in Section \ref{sec:conclusion}.

\section{Tenodesis Grasping Emulator}
\label{sec:design}

\begin{figure}[t]
    \centering
    \vspace{2mm}
    \includegraphics[width=1\linewidth]{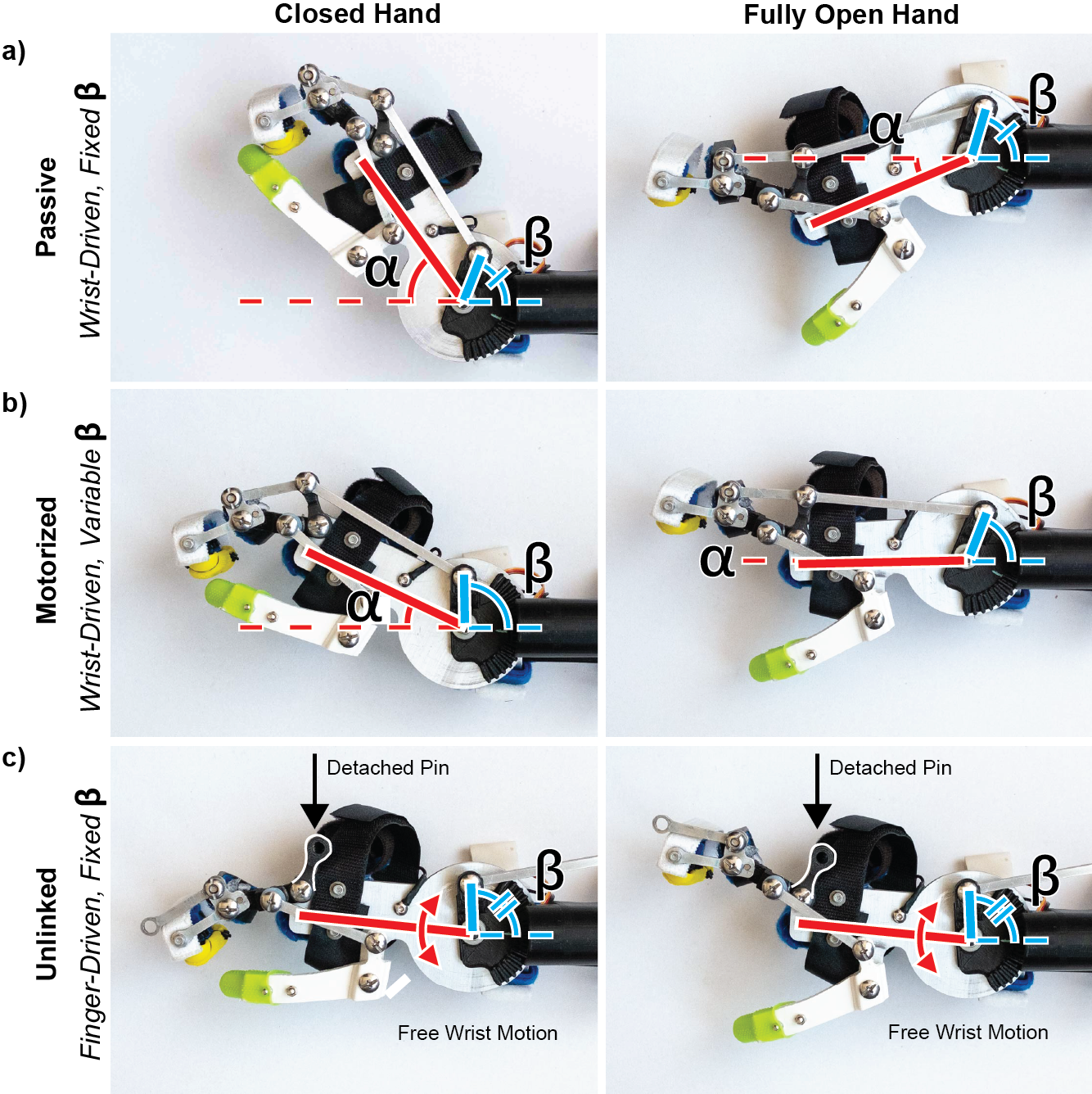}
    \caption{Passive, motorized, and unlinked device control modes demonstrating the wrist angles $\alpha$ and corresponding motor angles $\beta$ necessary to actuate the thumb for opening and closing the hand. (a) In passive mode, $\beta$ is fixed and large $\alpha$ actuates the thumb. (b) In motorized mode, thumb actuation requires smaller $\alpha$ due to assistance from $\beta$. (c) In unlinked mode, wrist motion is decoupled from finger position by disconnecting the linkage pin. $\beta$ is fixed and finger position actuates the thumb.}
    \vspace{-5mm}
    \label{fig:modes}
\end{figure}

To understand the impact of tenodesis grasping on upper body kinematics, we first design a device that mimics the technique by restricting some dexterity in normative subjects. The TGE is derived from the MWDO, which harnesses both body-powered and motorized actuation in parallel to create flexible device behaviors presented in our previous work \cite{McPherson2020Motor-AugmentedInjury}. The TGE constrains the four fingers to move together, allowing flexion and extension only, and restricts radial and ulnar deviation in the wrist.

\subsection{Artificial Thumb Actuation}
\label{sec:thumb}

Intuitive normative grasping contradicts the tenodesis grasping method, and such 
subjects may unintentionally apply large forces to the linkage by squeezing the thumb and fingers together. 
Especially in wrist-driven control,
this finger force 
counteracts the tenodesis grasp wrist motion and can backdrive the device motor, overcoming the grasping constraint. 
The TGE avoids this circumstance by replacing the MWDO thumb support with an artificial thumb (Fig. \ref{fig:testbed}a) actuated by wrist motion, to more reliably replicate the tenodesis grasping technique employed by individuals with SCI. As a result, contact with objects typically occurs with the device, rather than the skin of the subject; 
the TGE includes rubber fingertips to generate higher contact friction.

To wear the device with the artificial thumb, participants first don a nitrile glove on their right hand while keeping their thumb against the palm, without donning the thumb sheath. The fingers are then slipped into the loop near the distal interphalangeal joints and forearm straps are applied at the base. Additional support straps affixed to the palm prevent unintentional hand rotation within the device in the absence of the MWDO thumb support, as in Fig. \ref{fig:testbed}c.

\subsection{Modes of Operation}
\label{sec:modes}

When the motor is unpowered, the device acts like a standard passive WDO \cite{Portnova2018DesignInjury}, where wrist flexion and extension actuates the movement of the artificial thumb 
via two four-bar linkages in series; this \textit{passive mode} is pictured in Fig. \ref{fig:modes}a.
When the motor is activated, i.e. \textit{motorized mode}, both wrist and motor motions contribute to the grasp aperture. As a result, the motor manipulates how the device responds to wrist motion by measuring wrist angle with a potentiometer. Fig. \ref{fig:modes}b shows how motor assistance reduces the wrist angle required to actuate the thumb, compared to passive mode. 
Although more complex control strategies are possible, in this first study, the TGE applies a constant gain to reduce the wrist angle needed to actuate the thumb.

A final \textit{unlinked mode} is achieved by mechanically disconnecting one linkage pin, pictured in Fig. \ref{fig:modes}c, to decouple thumb actuation 
from the motor and wrist. We collect control data with the unlinked mode, 
where the subject can voluntarily, though jointly, flex and extend their fingers to actuate the thumb while still wearing the device. This control testing allows for isolating the effect of specific variables, since wearing the device could inherently influence behavior.

\section{Experimental Methods}
\label{sec:methods}

We ran a human subjects trial with the TGE 
to (1) measure the compensatory motions associated with WDO tenodesis grasp constraint (passive mode), as compared with normative function (unlinked mode), and (2) test whether, and under what conditions, motor assistance (motorized mode) performs better or worse than WDO function (passive mode).


\subsection{Participants and Experimental Procedure}
\begin{figure}[t]
    \centering
    \vspace{2mm}
    \includegraphics[width=1\linewidth]{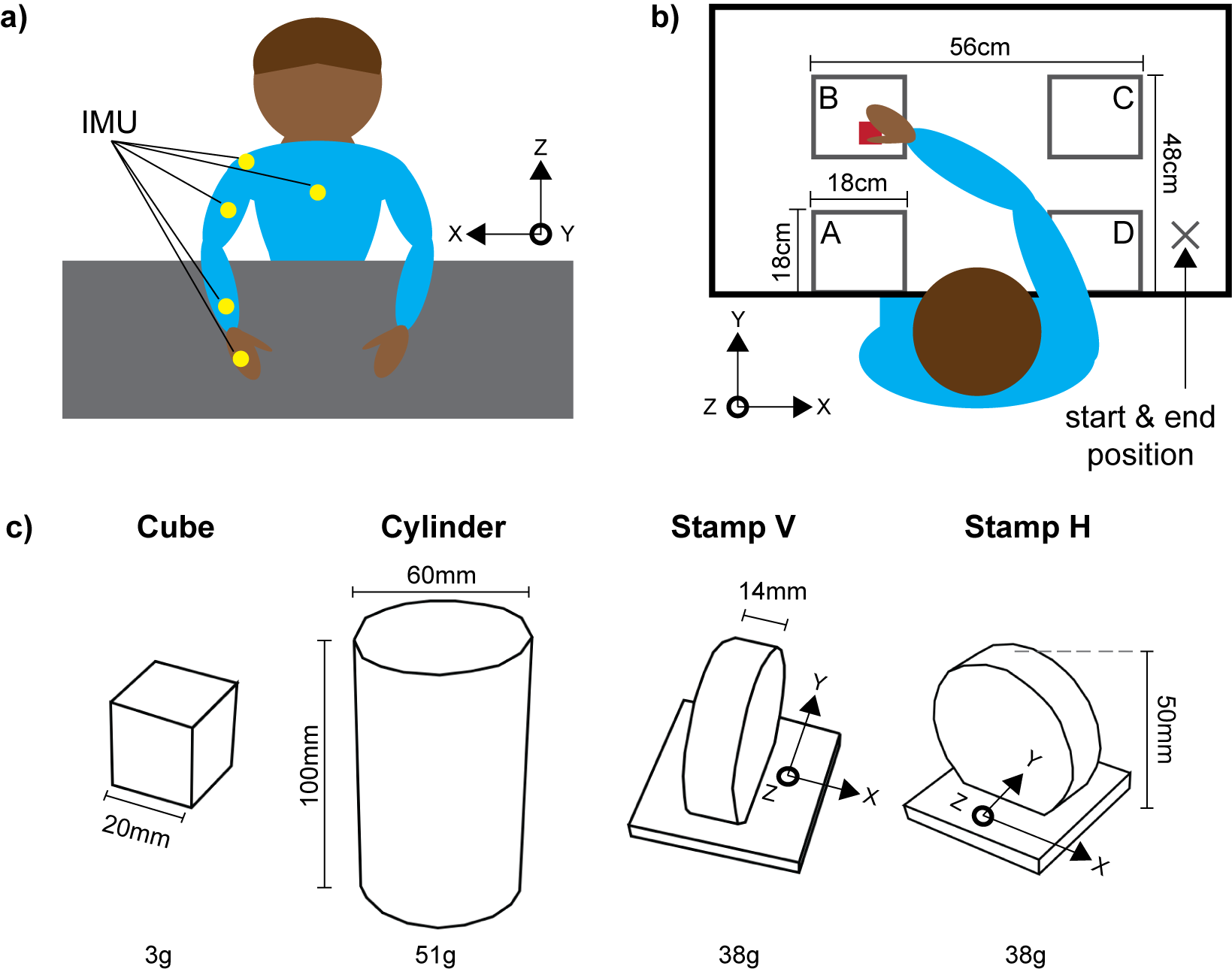}
    \caption{(a) IMU marker locations (yellow) worn during trials. (b) Four GRT locations  marked A,B,C,D and grasping workspace setup. (c) Objects used in grasping trials, including cube, cylinder, and stamp in
    two orientations: handle positioned perpendicular to the subject's body (Stamp V) and handle positioned parallel to the subject's body (Stamp H). 
    }
    \vspace{-3mm}
    \label{fig:objloc}
\end{figure}

Under the UC Berkeley IRB-approved protocol \#2020-02-12983, 6 subjects (3 females and 3 males, aged 23.6 $\pm$ 2.3 years old) with unimpaired hand function who had never been diagnosed with a neurological or motor disorder participated in the study. Participants performed an assortment of seated pick-and-place tasks in a modified Grasp and Release Test (GRT), while wearing the TGE on their right hand in each of the three modes introduced in Section \ref{sec:modes}. Throughout all GRT tasks, videos of the upper body were recorded while kinematic data was collected from 5 IMU sensors (Xsens DOT) attached to the torso, shoulder, upper arm, forearm, and hand, corresponding to the TGE-donned limb (Fig. \ref{fig:objloc}a). Each 9-axis sensor ran a proprietary sensor fusion algorithm to compensate for drift.

The trials consisted of grasping an object from one of four locations and releasing it in a second location, while operating the TGE in one of the three modes. Each task began and ended with the subject's right hand resting on the marked position, indicated in Fig. \ref{fig:objloc}b. 
Participants sat at a fixed-height stool to allow unconstrained torso movement in all directions and received instructions to perform tasks at a comfortable and natural pace, without time limitations.

Subjects performed the GRT by manipulating three test objects,
selected to include both power wrap and precision pinch grasps. Of the three test objects, we selected two
axially symmetric objects (cube, cylinder) and one 
asymmetric object (stamp) to evaluate the impact of constraining the hand position during grasp and release events. The cube and cylinder could be grasped and released in any upright orientation, however subjects were instructed to grasp and release the stamp with the handle in either a consistent vertical (Stamp V) or horizontal (Stamp H) orientation, which were considered as separate objects (and hereinafter referred to as such) during analysis (Fig. \ref{fig:objloc}c).
Dropped objects and unsuccessful grasps were
noted and excluded from kinematic analysis. For each trial combination of mode, object, and location (going from one region to another, in one direction), no more than 2 trials were removed for each subject.

TGE mode was randomized for each subject and all trials for a single mode were completed before switching to another mode. Within each mode, object ordering was also randomized. Subjects grasped and released each object 
in a ``forward'' (A-C, C-D, D-B, B-A) and ``reverse'' (A-B, B-D, D-C, C-A) order, directed on a screen in front of them. As a result, 4 trials were gathered for each movement direction, for 32 trials of each object and mode combination, and 384 trials per subject total.  
Between modes, participants received unstructured practice time with all of the objects to become accustomed to each mode before performing the trials.

In addition to performing the GRT tasks in each mode, we surveyed each participant to evaluate their perceived difficulty of the tasks they performed and modes they experienced. The subjects were asked to (1) rank the objects in order of overall difficulty, and (2) identify the most difficult location for each object and mode combination.

\subsection{Compensation Metrics and Data Analysis}

\begin{figure}[t]
    \centering
    \vspace{2mm}
    \includegraphics[width=1\linewidth]{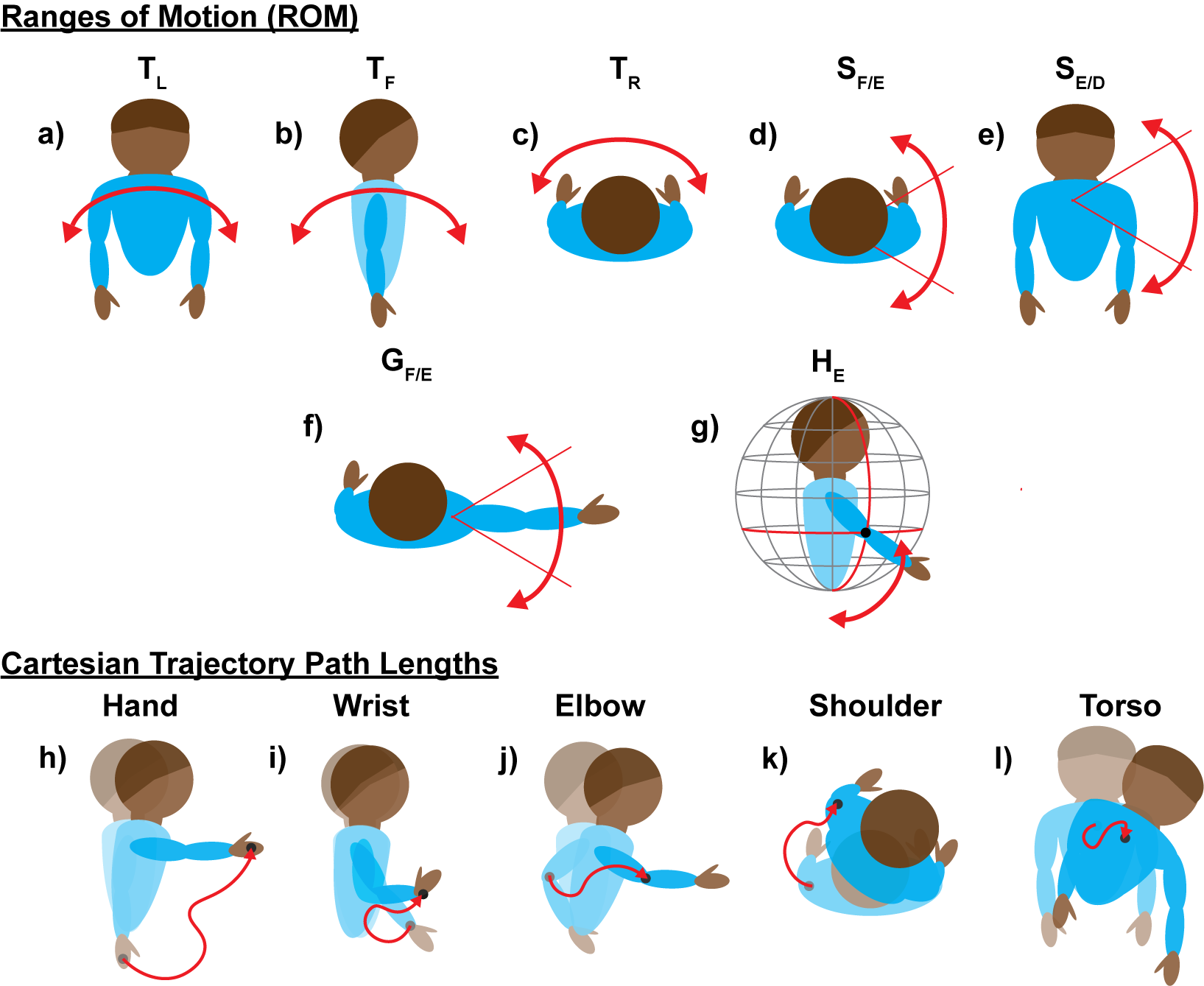}
    \caption{Body angles measured for range of motion (ROM) metrics: (a)-(c) trunk movements in the lateral, forward, and rotational directions, (d)-(e) shoulder girdle flexion/extension and elevation/depression, (f) glenohumeral joint horizontal flexion/extension, (g) humeral elevation. Body points tracked to determine Cartesian trajectory length, calculated as the sum of Euclidean distances between each point throughout the trial: (h) hand position, (i) wrist joint position, (j) elbow joint position, (k) shoulder joint position, (l) torso position.}
    \label{fig:bodyangsandsegs}
    \vspace{-3mm}
\end{figure}

We parametrize motion kinematics in two ways, using the motion capture data collected, as metrics for compensatory behavior: (1) the range of motion (ROM) of upper body joint angles, and (2) the total Cartesian trajectory length traveled by points on the upper body. 
The ROM metric additionally reduces the negative impact of the drift we saw in the sensor data by avoiding absolute joint measurements in analysis.

These two metrics provide different perspectives, by looking at both the maximum effects of an individual joint and the aggregate effect of multiple joints acting in unison over time. While ROM has been established and utilized by different groups as a practical quantifier of compensation \cite{Montagnani2015IsProstheses,Murgia2010TheLiving}, complementing this metric with trajectory lengths provides motion information not captured using ROM alone \cite{Spiers2018ExaminingWorkspace,Gloumakov2020TrajectoryStudy}.

\begin{figure}[t]
    \centering
    \vspace{2mm}
    \includegraphics[width=1\linewidth]{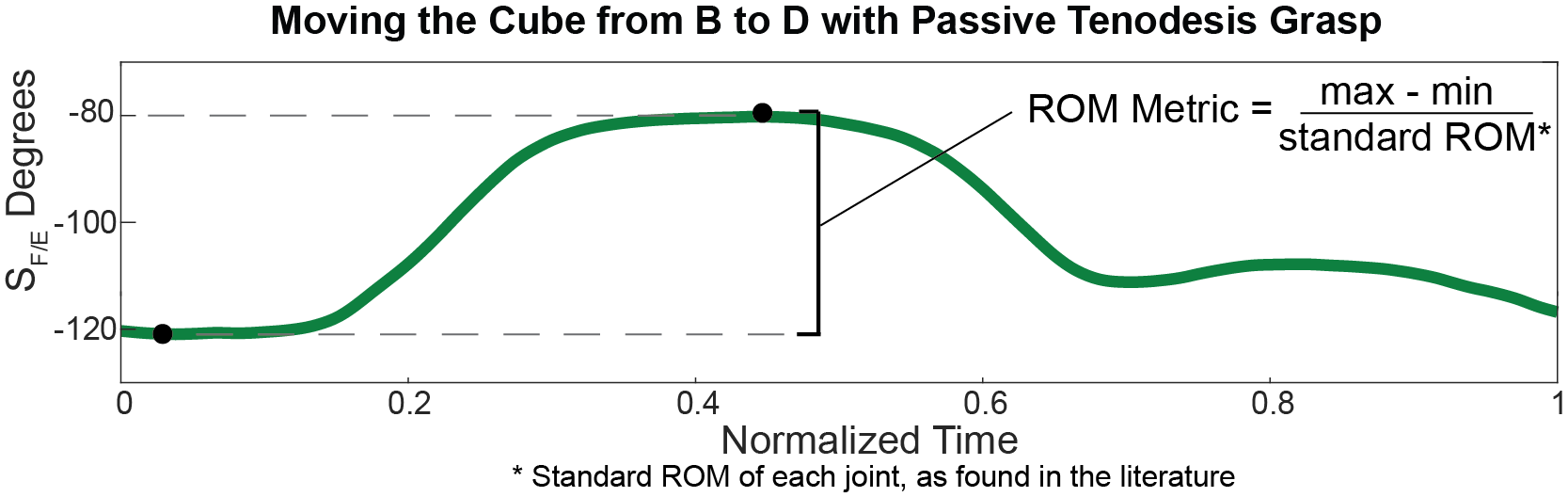}
    \caption{Sample calculation of ROM for a single subject's shoulder flexion/extension while moving the cube from location B to D and using passive tenodesis grasping.}
    \label{fig:ROMmetric}
\end{figure}

ROM metrics utilize body angles identified in \cite{Montagnani2015IsProstheses, Murgia2010TheLiving,Doorenbosch2003TheMovements}, and shown in Fig. \ref{fig:bodyangsandsegs}a-g. These include lateral, forward, and rotational trunk angle (T$_L$, T$_F$, T$_R$), flexion/extension and elevation/depression of the shoulder girdle (S$_{F/E}$, S$_{E/D}$), horizontal flexion/extension of the glenohumeral joint (G$_{F/E}$), and humeral elevation (H$_{E}$). Selected point trajectories include those of the end of the hand, the top of the torso, and the wrist, elbow, and shoulder (glenohumeral) joints, as shown in Fig. \ref{fig:bodyangsandsegs}h-l.
ROM is the difference between the maximum and minimum body angle in a single task, normalized by the standard ROM for that joint \cite{Neumann2013KinesiologyRehabilitation,I.A.Kapandji1970TheLimb,Gloumakov2020DimensionalityOrientation}.
The ROM metric may not capture the full motion, as shown in Fig. \ref{fig:ROMmetric}, where it fails to consider 
the second smaller peak in S$_{F/E}$ movement that occurs toward the end of the trial. The trajectory length metric of the shoulder joint can account for this additional movement.

Cartesian trajectory length is the total Euclidean distance traveled by a point on the body throughout a single task. Forward kinematics determine the location of body point positions with respect to each subject's body dimensions.

Analysis of variance (ANOVA) is used to compare the compensation metrics across the primary factors in each task: TGE mode, object, and location. For the collected dataset, the Shapiro-Wilks test indicated about 61\% of the distributions were normal and Levene's test indicated that about 85\% of groups were homoscedastic. Tukey’s honest significant difference criterion is used in post-hoc pairwise comparison tests 
between factor variables.


\section{Results}
\label{sec:results}

\begin{figure}[t]
    \centering
    \includegraphics[width=1\linewidth]{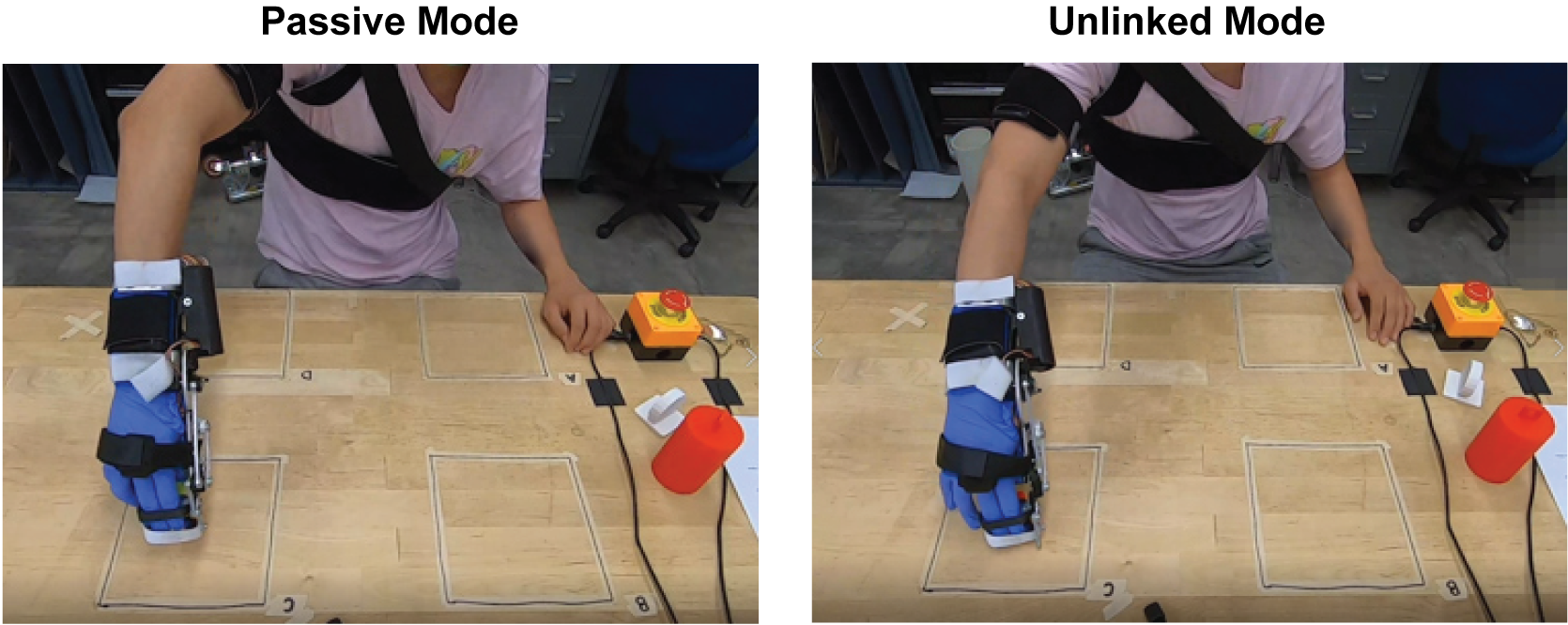}
    \caption{Subject grasping the cube at location C in passive and unlinked modes.}
    \label{fig:compensation}
\end{figure}

\begin{figure*}[t]
    \centering
    \vspace{2mm}
    \includegraphics[width=0.8\linewidth]{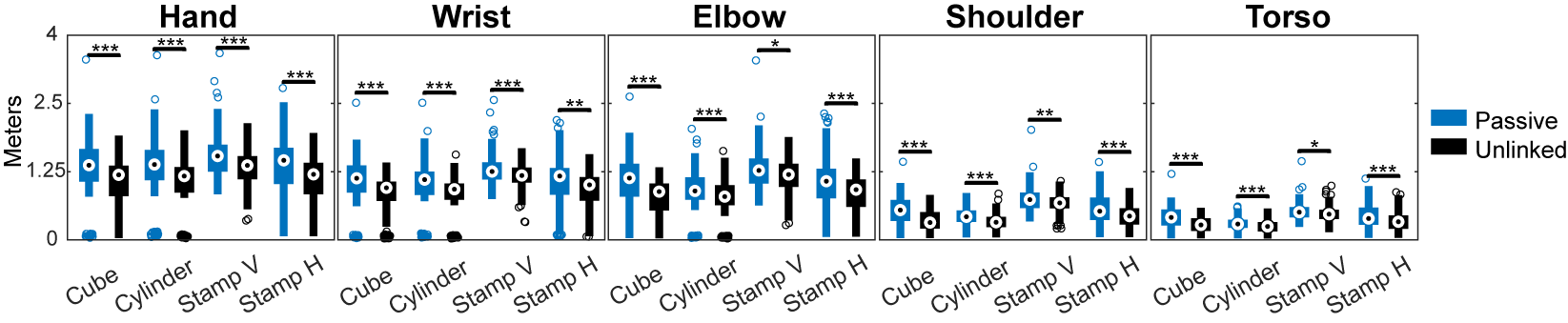}
    \vspace{-3mm}
    \caption{Distribution of Cartesian trajectory lengths, in meters, of passive tenodesis grasp (blue) and unlinked (black) mode for four objects and five body points. Box and whiskers plot represents median values by encircled black dots within a rectangular box denoting the interquartile range and flanking whiskers. ***p$<$0.001, **p$<$0.01, *p$<$0.05.}
    \label{fig:tenovcontrol_box}
\end{figure*}
\subsection{Constraints Imposed by Tenodesis Grasping}
\label{sec:resultsPassive}

Subjects exhibit a wide range of compensatory motions, when comparing the passive and unlinked modes of the TGE, one example which can be seen in Fig. \ref{fig:compensation}. A multi-factor ANOVA shows that each of the factors (mode, object, and location) is always statistically significant (p$<$0.05) when accounting for differences in the other two factors, respectively. Thus, we first compare the passive tenodesis grasp and unlinked TGE configurations by object, across all locations.

\begin{figure}[t]
    \centering
    \includegraphics[width=0.8\linewidth]{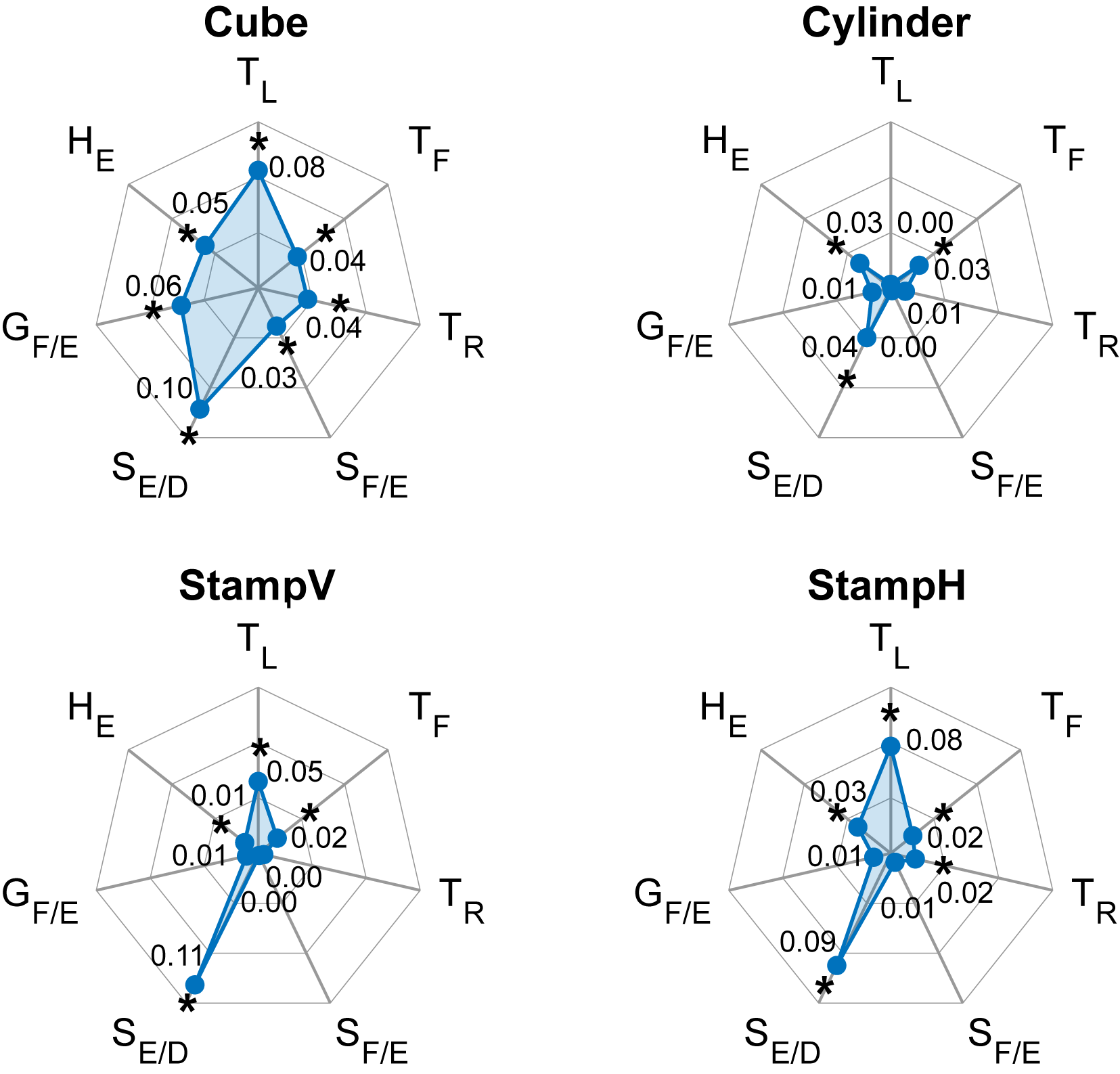}
    \vspace{-3mm}
    \caption{Relative compensatory ROM (as calculated by the equation in Fig. \ref{fig:ROMmetric}) when using passive tenodesis grasp, compared to unlinked. 0.00 indicates passive mode induces the same amount of movement as unlinked, or no compensation, for four objects and seven body angles. *p$<$0.05.}
    \label{fig:tenovcontrol_spider}
\end{figure}

Kinematic analysis shows that all upper body points travel significantly longer trajectory lengths for all objects when using passive tenodesis grasp (Fig. \ref{fig:tenovcontrol_box}). The median path lengths for Stamp V are consistently longer than those of the other three objects. For most body joints, the ROM exhibited in the passive mode is consistently larger than the unlinked mode, with the exception of T$_L$ and S$_{F/E}$ when moving the cylinder, and T$_R$ and S$_{F/E}$ when moving Stamp V, where no difference is exhibited; Fig. \ref{fig:tenovcontrol_spider} shows the differences between these modes. For every object, subjects exhibit larger ROM with passive tenodesis grasp that are statistically significant (p$<$0.05) for the T$_F$, S$_{E/D}$, and H$_E$ angles. When manipulating the cube, participants show significant differences in ROM between modes for all seven body angles. Participants also have more difficulty maneuvering the cube (either failing to pick up or dropping it) than any other object. The two stamps induce a similar pattern of ROM, with the most prominent difference between modes arising from T$_L$ and S$_{E/D}$. While Stamp H shows larger relative compensatory ROM for most body angles than compared to Stamp V, participants drop Stamp V nearly twice as often as they drop Stamp H. Compared to the other objects, the cylinder shows relatively smaller compensatory ROM for the seven body angles. 

\begin{figure}[t]
    \centering
    \includegraphics[width=0.85\linewidth]{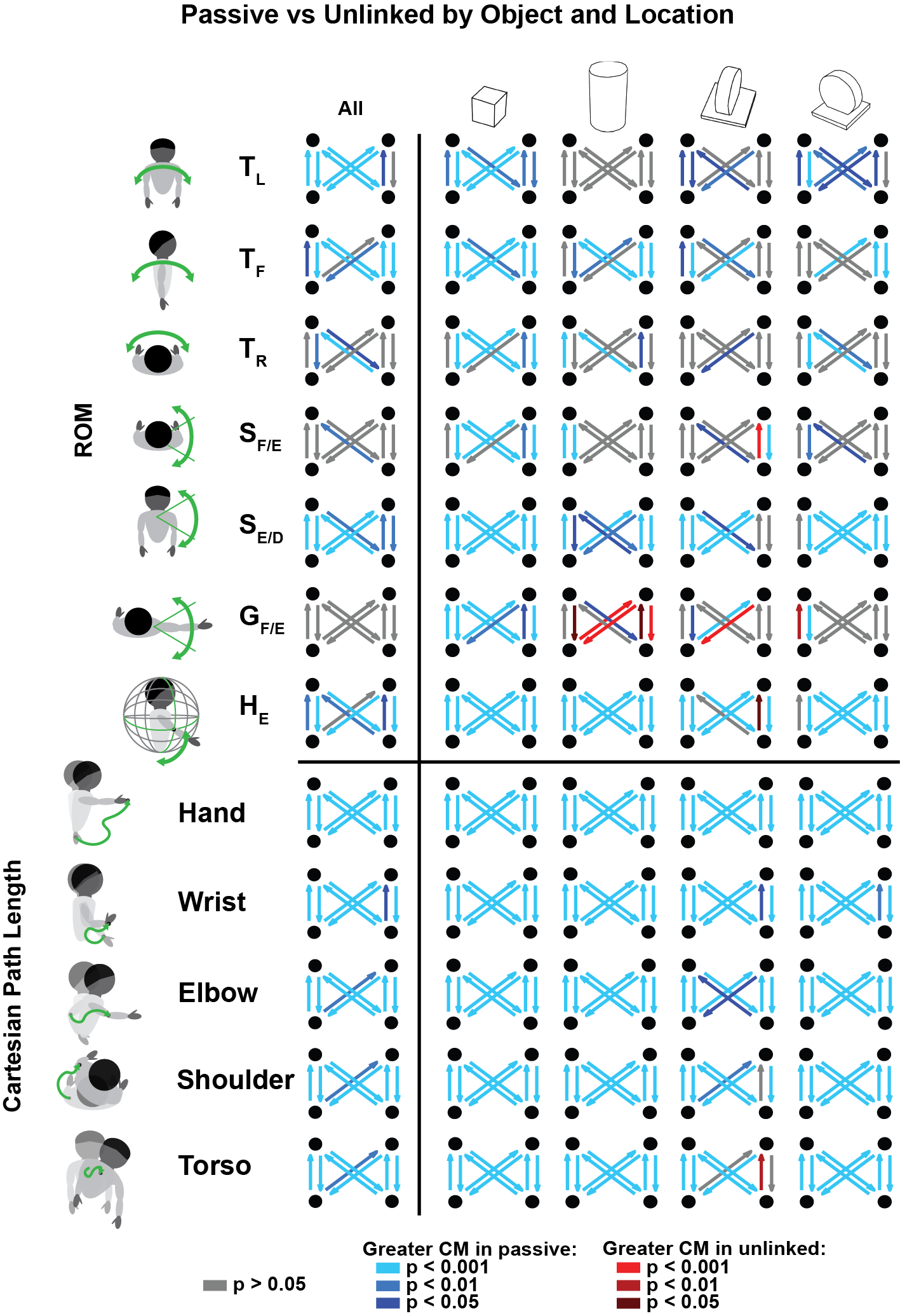}
    \vspace{-2mm}
    \caption{Comparison of relative compensation between passive and unlinked modes, measured by angular ROM and Cartesian trajectory length for each object. Blue arrows indicate that passive yields greater compensatory movement (CM), red arrows indicate that unlinked yields greater CM, and gray arrows indicate no significant difference. Black dots represent the locations pictured in Fig. \ref{fig:objloc}b and the arrows denote the direction of grasp and release between two locations.}
    \label{fig:tenovcontrol_x}
\end{figure}

Since the multi-factor ANOVA indicates that location is a statistically significant factor, we also explore the impact of location on the compensation metrics. Fig. \ref{fig:tenovcontrol_x} depicts the grasping workspace locations (represented by black dots) and the eight directions in which the objects are manipulated (represented by colored arrows) for each compensation metric and object. As shown in Fig. \ref{fig:tenovcontrol_x}, the most common case across all conditions and metrics is that the passive mode produces more compensatory movement than unlinked. Across all objects (leftmost column), the differences in trajectory lengths of all body points are statistically significant for all locations, and each body point travels farther in the passive tenodesis grasp configuration than unlinked configuration. In the same column, larger compensation with tenodesis grasp during grasp and release from location D to B is statistically significant for all body angles, except for G$_{F/E}$.

As illustrated in columns 2-5 of Fig. \ref{fig:tenovcontrol_x}, the pattern of relative trajectory lengths between passive tenodesis grasp and the unlinked configuration for each location represents the same pattern as seen with the cube, cylinder, and Stamp H alone. With Stamp V, however, the shoulder and torso during passive tenodesis grasp travel slightly less than or a similar distance to the unlinked configuration when moving between locations C and D. We observe a similar trend in the ROM for the Stamp V. Each of the seven body angles between these two locations yields instances of insignificant difference or less motion in passive tenodesis grasp than unlinked. 

\subsection{Comparing Motorized and Passive Tenodesis Grasping}
\label{sec:resultsMotor}

\begin{figure}[t]
    \centering
    \vspace{2mm}
    \includegraphics[width=0.85\linewidth]{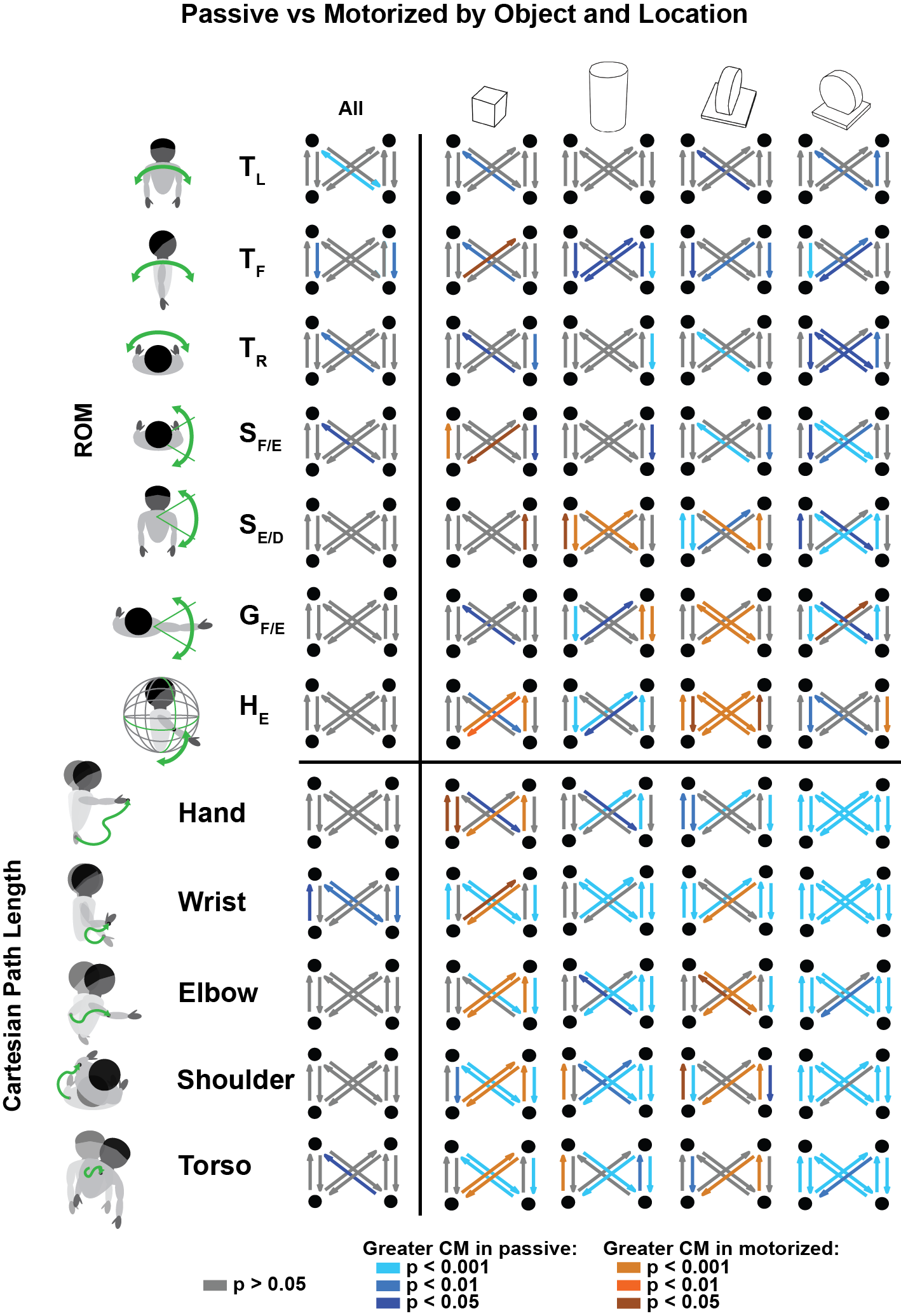}
    \vspace{-3mm}
    \caption{Comparison of relative compensation between passive and motorized modes, measured by angular ROM and Cartesian trajectory length for each object. Blue arrows indicate that passive yields greater compensatory movement (CM), orange arrows indicate that motorized yields greater CM, and gray arrows indicate no significant difference. Black dots represent the locations pictured in Fig. \ref{fig:objloc}b and the arrows denote the direction of grasp and release between two locations.}
    \label{fig:passivevaugment_x}
\end{figure}

A multi-factor ANOVA on mode (passive and motorized), object, and location shows a  statistically significant (p$<$0.05) effect of mode on compensation 
when accounting for differences in object and location for the all three trunk angles, and all five body points' trajectory lengths. Both object and location are statistically significant for all body angles and body points. 
As shown in the leftmost column of Fig. \ref{fig:passivevaugment_x}, for all objects, few location directions show a statistically significant difference in compensation between TGE modes across either metric. However, consistently moving objects from D to B show more ROM compensation in passive mode for three body angles (T$_L$, T$_R$, S$_{F/E}$) and trajectory length compensation for two body points (wrist and torso).

Comparing the body- and motor-powered TGE modes by both the location and object in columns 2-5 of Fig. \ref{fig:passivevaugment_x}, the cube and Stamp V show more instances of greater compensation resulting from motorized mode, over half of which occur when moving the objects between diagonally-spaced locations. Moving the cube between locations A and C induces longer trajectory lengths in motorized mode for all body points, but moving between the opposite diagonal locations, B and D, induces more compensatory movement in passive mode. The cylinder and Stamp H, however, show predominantly more instances of larger compensation in passive mode than motorized mode.
Participants also show improvement in the number of failed attempts to maneuver the cube and Stamp H from passive to motorized mode, while the number of failed attempts between modes is similar for the cylinder and Stamp V.



\subsection{Subject Survey}


Participants predominantly (85\%) report Stamp V is the most difficult object to maneuver and the cylinder is the easiest object to maneuver (66\%). Of the trials with Stamp V, all participants additionally report that grasping is most difficult when starting at location D. Both C and D are also highly reported as difficult grasping locations across all objects, and every subject reports difficulty starting from those locations for at least one object in passive mode. When asked about areas of discomfort following the experiment, 66\% of subjects report fatigue in the shoulder.

\section{Discussion}
\label{sec:discussion}

As seen in Section \ref{sec:resultsPassive}, the tenodesis grasping technique 
usually generates body compensation, or atypical movements. 
This is of particular interest for people with C6-7 SCI, who commonly report injuries and pain in the shoulder 
as a result of overuse and stress on the rotator cuff \cite{Apple2004ChapterService,Sie1992UpperPatient}; tenodesis grasping may contribute to increased biomechanical forces in this region. 
The effect of the wrist-driven constraint on task difficulty varies across different objects and reaching directions, as indicated in both the ROM measures and surveys. This may help to explain why people with limited hand dexterity perform non-prehensile pushing movements, when able, to move objects into a better location and orientation in their work-space prior to grasping \cite{Cochran2018AnalyzingAmputees}.




As highlighted in Section \ref{sec:resultsMotor}, the TGE control mode that reduces more compensation
depends on object type and location. Thus,  
a dynamic control system that alters mode, or wrist-finger gain, in response to the environment and body pose has the potential to ease body compensation in wrist-driven systems. 
Some versions of the WDO include manual clutches or brakes to enable such flexibility \cite{bacon1978sequential}, however, 
they require bimanual dexterity or assistance of a fitting technician to change the device functionality. Hybrid motorized controllers, such as with the MWDO and TGE, instead offer more convenient versatility in dynamic environments.

\begin{figure}[t]
    \centering
    \vspace{2mm}
    \includegraphics[width=1\linewidth]{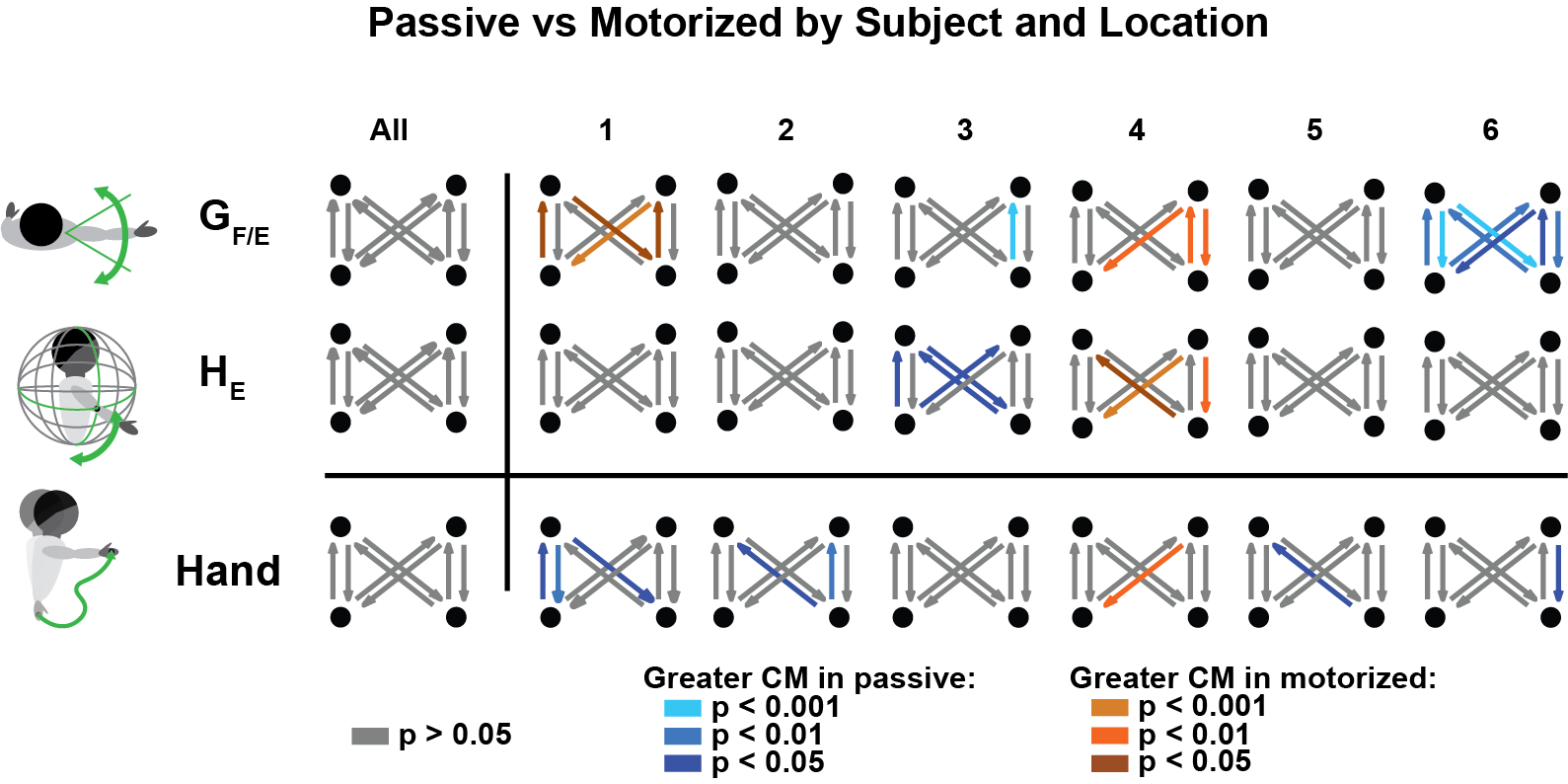}
    \vspace{-6mm}
    \caption{Relative compensatory movement (CM) between motorized and passive tenodesis grasp configurations, measured by angular ROM and Cartesian trajectory length for each subject. Blue arrows indicate that passive  yields  greater CM, orange arrows indicate that motorized yields greater CM, and gray arrows indicate no significant difference. Black dots represent the locations pictured in Fig. \ref{fig:objloc}b and the arrows denote the direction of grasp and release between two locations.}
    \label{fig:bySubject}
    \vspace{-4mm}
\end{figure}

Fig. \ref{fig:bySubject} provides a partial look at how the comparison between the passive and motorized modes separates between the 6 individual subjects, showing that different subjects respond differently to the device modes. These differences may result from the order that the subject was exposed to the different device modes, learning effects, or differences in personal preference or body dimensions. Future work will decode these potential effects, in order to understand whether subjects benefit from a one-size-fits-all control scheme or a more personalized option.


\section{Conclusion}
\label{sec:conclusion}

People with C6-7 SCI use a grasping strategy that constrains manual dexterity by coupling the closing and opening of the hand to wrist flexion and extension.
The TGE is a novel adaptor device for evaluating the effect of tenodesis grasping on upper-body kinematics in unimpaired subjects. Our results indicate that tenodesis grasp leads to significant amounts of atypical movement in the upper body, that may ultimately affect the utility and comfort of such wrist-driven orthotics. 
Introducing motorized assistance in a wrist-driven system can ease the severity of some of these compensatory movements. This work aims to guide the control design of such hybrid-actuated devices that maintain the benefits of body-powered actuation, while allowing for functional versatility across different user preferences and tasks.


Compensation, and the role of tenodesis grasping, is likely different between unaffected subjects and subjects with SCI who are already familiar with the strategy. 
Following this initial study with the TGE, we plan to conduct subsequent studies with the MWDO in subjects with SCI to evaluate the accuracy of emulating tenodesis grasping in individuals without SCI. Future studies will also shift the focus more toward ADLs, to represent more realistic scenarios.

\section*{Acknowledgment}
\thanks{This work is supported by the Hellman Fellows Fund and the University of California, Berkeley. Erin Chang is additionally supported by the National Science Foundation Graduate Research Fellowship Program under Grant No. DGE 2146752. Any opinions, findings, and conclusions or recommendations expressed in this material are those of the authors and do not necessarily reflect the views of the funding agencies. The authors acknowledge the support of the members of the Embodied Dexterity Group.}

\bibliographystyle{IEEEtran}
\bibliography{MWDO}

\end{document}